\documentclass[letterpaper, 10 pt, conference]{ieeeconf}  % Comment this line out if you need a4paper

\IEEEoverridecommandlockouts                              % This command is only needed if 
\usepackage{todonotes}

\usepackage{bm}
\usepackage{amsmath}
\usepackage{amsfonts}
 \usepackage{algorithm}
\usepackage{algorithmic}
\usepackage{blindtext}
\usepackage{scalefnt}

\newcommand{\onum}[1]{\overline{{#1}}} 	%upper limit interval
\newcommand{\unum}[1]{\underline{{#1}}}  	%lower limit interval
\graphicspath{{img/}{.}}

\title{\LARGE \bf
On-line feasible wrench polytope evaluation based on human musculoskeletal models: an iterative convex hull method
}

\author{Antun Skuric$^{1}$, Vincent Padois$^{1}$, Nasser Rezzoug$^{1,2}$, David Daney$^{1}$ % <-this % stops a space
\thanks{$^{1}$  Auctus,  Inria  /  IMS  (Univ.  Bordeaux /  Bordeaux  INP  /  CNRS  UMR  5218),  33405  Talence,  France \textit{email: firstname.lastname@inria.fr}}
\thanks{$^{2}$University of Toulon, Toulon, France  \textit{email:rezzoug@univ-tln.fr}}%
\thanks{Human-Robot experiments including direct physical interactions in the 
Auctus research team have been performed in accordance with Inria
Ethics committee recommendations.}
}

\begin{document}
\scalefont{0.9825}

\maketitle
\thispagestyle{empty}
\pagestyle{empty}

%%%%%%%%%%%%%%%%%%%%%%%%%%%%%%%%%%%%%%%%%%%%%%%%%%%%%%%%%%%%%%%%%%%%%%%%%%%%%%%%
\begin{abstract}
Many recent human-robot collaboration strategies, such as \textit{Assist-As-Needed} (AAN), are promoting human-centered robot control, where the robot continuously adapts its assistance level based on the real-time need of its human counterpart.
One of the fundamental assumptions of these approaches is the ability to measure or estimate the physical capacity of humans in real-time.

In this work, we propose an algorithm for the feasibility set analysis of a generic class of linear algebra problems. This novel iterative convex-hull method is applied to the determination of the feasible Cartesian wrench polytope associated to a musculoskeletal model of the human upper limb. The method is capable of running in real-time and allows the user to define the desired estimation accuracy.  
The algorithm performance analysis shows that the execution time has near-linear relationship to the considered number of muscles, as opposed to the exponential relationship of the conventional methods. Finally, real-time robot control application of the algorithm is demonstrated in a \textit{Collaborative carrying} experiment, where a human operator and a \textit{Franka Emika Panda} robot jointly carry a 7kg object. The robot is controlled in accordance to the AAN paradigm maintaining the load carried by the human operator at 30\% of its carrying capacity.

\end{abstract}

%%%%%%%%%%%%%%%%%%%%%%%%%%%%%%%%%%%%%%%%%%%%%%%%%%%%%%%%%%%%%%%%%%%%%%%%%%%%%%%%
\section{Introduction}

The number of collaborative robots, robots designed to closely coexist with humans, is increasing exponentially in recent years \cite{ajoudani2018progress}, paving the way for more comprehensive and human-centered solutions for human robot collaboration. However, due to the instantaneous nature of the physical interaction, it becomes increasingly challenging to design and tune control laws for the collaborative systems \textit{a priori}. Accounting, in advance, for the high degree of variability induced by humans usually results in very conservative safety limits and reduces the overall collaboration efficiency considerably. Therefore, new on-line capable and accurate metrics are needed to improve performance of the collaborative systems while ensuring safety for all humans involved. \textit{Assist-As-Needed} (AAN)\cite{carmichael2013admittance} is one recent example of such paradigm, where a robot continuously adapts the assistance level taking in consideration the capacity of the human operator in real-time.

Several capacity metrics have been developed over the years, first by the robotics community\cite{yoshikawa1985manipulability}\cite{chiacchio_evaluation_1996}, and later adapted to the human motion analysis\cite{sasaki2011vertex}\cite{khatib2009robotics}\cite{biomechanics1010008}. The most well known representatives being wrench, twist and acceleration capacity. While twist and acceleration capacity represent the level of dexterity in different directions, wrench capacity represents the human ability to apply and resist external wrenches in arbitrary directions, both being very important safety-wise and performance-wise. 

\begin{figure}[!t]
    \centering
    \includegraphics[width=0.9\linewidth]{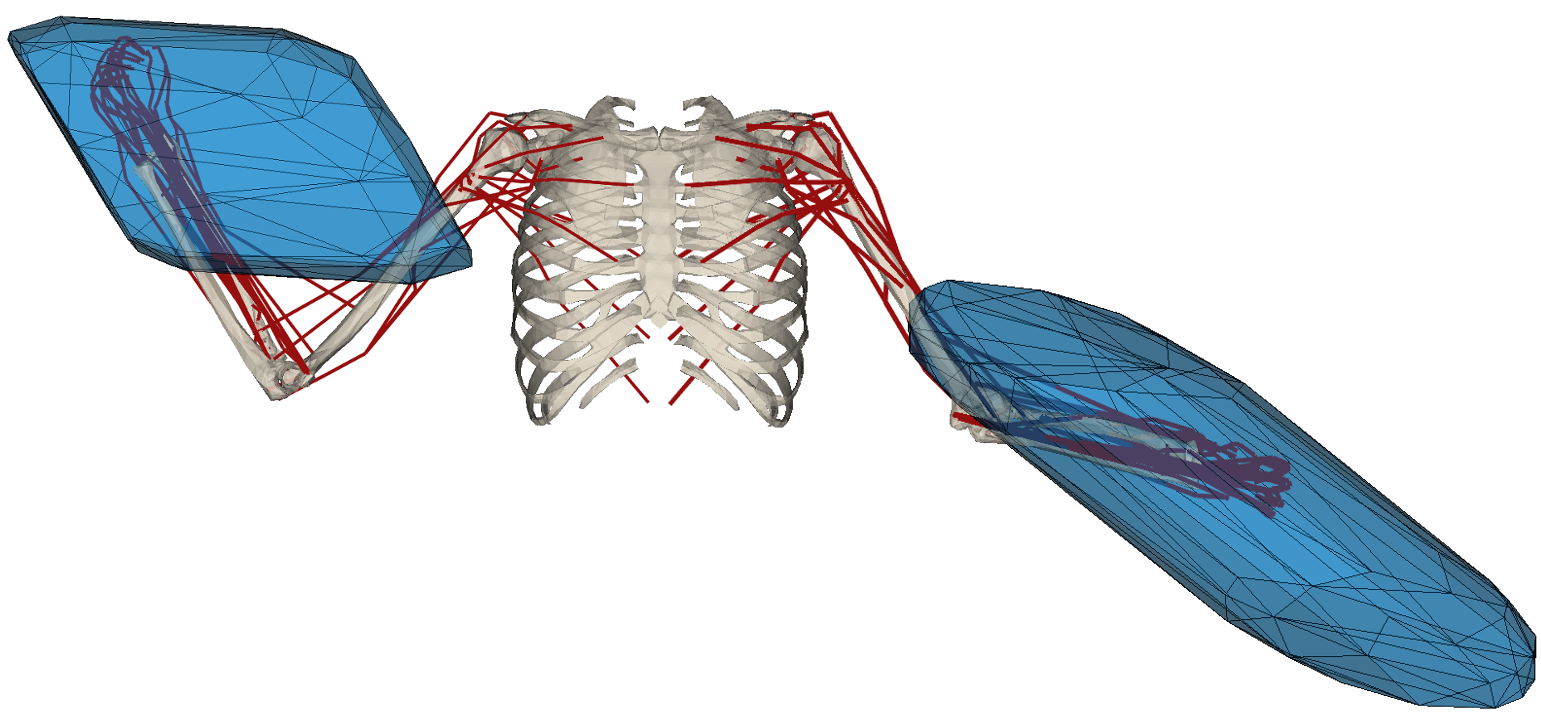}
    \caption{Cartesian force polytope of a musculoskeletal model of both human upper limbs  \cite{saul2015benchmarking} with 7Dof and 50 muscles each, visualised in \textit{pyomeca bioviz} \cite{Michaud2021}. The polytopes are scaled with a ratio 1m : 1000N}
    \label{fig:polytope_showoff}
\end{figure}

The human body capacities are governed by many different factors \cite{nasa}. Many of them are physiological and biomechanical factors, such as muscle strength, bone length and level of fatigue. The others are purely kinematic such as posture, joint velocities and accelerations. 
Among the most complete human body modeling approaches are the musculoskeletal models, describing the relationship between the activation of human muscles and the body dynamics. 
Over the years, many, more or less complete, models have been developed and experimentally validated \cite{holzbaur2005model}\cite{WU20163626}\cite{rajagopal2016fullbody}, as well as different open-source software platforms created for their analysis \cite{opensim}\cite{Michaud2021}. 

The exact wrench, twist and acceleration capacity of the musculoskeletal model can be represented in a form of convex polytopes. However due to the relatively high number of muscles usually considered, the exact evaluation of these polytopes is a time consuming process. Therefore, in order to reduce the complexity of their calculation, apart from using less complex models, various approximate techniques have been proposed, using machine learning \cite{hernandez2018force}, ellipsoids \cite{petric2019assistive} and approximate polytope estimation techniques \cite{carmichael2011Towards}\cite{chen2018strength}.  However, all of these approaches reduce the precision and confidence in the calculated human capacity. 

In this paper, a new algorithm for polytope evaluation is proposed. It allows for the efficient evaluation of the feasible wrench polytope of musculoskeletal human models, given a user specified accuracy. The efficiency is characterized by the low and near-linear complexity with respect to the number of muscles and accuracy. This allows for an online update of the polytope at a rate of 100 ms for 32 muscles and an accuracy of 5 N (200 ms for 50 muscles). Beyond these application specific characteristics, the proposed algorithm is generic to a class of implicitly defined linear algebra feasibility problems
\begin{equation}
    A\bm{x} = B\bm{y},\quad  \bm{y}\in \left[\unum{\bm{y}}, \onum{\bm{y}}\right]
    \label{eq:eq_general}
\end{equation}
where 
\begin{equation}
   \bm{x} \in \mathbb{R}^m, ~\bm{y} \in \mathbb{R}^d, ~A\in\mathbb{R}^{n \times m}, ~B\in\mathbb{R}^{n \times d}, ~ d\geq n \!\geq\! m
\end{equation}
In robotics, this class of problems can be found in many different domains such as when calculating feasible twist capacity of object grasping \cite{Prattichizzo2016}, feasible twist capacity of collaborating robots forming a closed chain \cite{bicchi2000manipulability}, or feasible wrench capacity of multilink cable driven parallel robots \cite{lau2016caspr}\cite{sheng2020operational}. %In the context of this paper, the proposed algorithm is used for feasible wrench polytope evaluation of musculoskeletal models. 
%Additionally, similar analysis can be done for the velocity and acceleration capacity as well.

Section \ref{ch:problem_formualtion} recapitulates the mathematical formulation of the feasible wrench polytope for musculoskeletal models, followed by the description of the proposed algorithm in Section \ref{ch:algorihtm}. In Section \ref{ch:results} the performance characteristics of the algorithm are analysed and compared with state of the art methods. Finally, a demonstrative experiment is shown where the proposed algorithm is used for real-time collaborative robot control.

\section{Feasible wrench polytopes formulation}
\label{ch:problem_formualtion}
The dynamical equation of a general musculoskeletal system expressed in joint space can be formulated as
\begin{equation}
    \underbrace{M(\bm{q})\ddot{\bm{q}} + C(\bm{q},\dot{\bm{q}})\dot{\bm{q}}}_{\bm{\tau}_d} + \bm{\tau}_g(\bm{q}) = \bm{\tau} - J^{T}(\bm{q})\bm{f}  
    \label{eq:human_dynamics}
\end{equation}
where $\bm{q},\dot{\bm{q}},\ddot{\bm{q}} \in \mathbb{R}^n $ are the joint level generalised coordinates, velocities and accelerations. $M$ is the inertia matrix, $C$ is the Coriolis-centrifugal matrix, $\bm{\tau}_g$ is the joint torque produced by gravity, $\bm{\tau}$ is the joint torque vector generated by the muscle-tendon units, $J$ is the end effector Jacobian matrix and $\bm{f} \in \mathbb{R}^m$ is an $m$-dimensional external Cartesian wrench vector. $\bm{\tau}_d$ represents the joint torques induced by the body motion and  kineto-static conditions $\dot{\bm{q}}\!=\!\ddot{\bm{q}}\!=\!0$ yield $\bm{\tau}_d = 0$.  
\begin{equation}
    \bm{\tau} - \bm{\tau}_d(\bm{q},\dot{\bm{q}},\ddot{\bm{q}}) - \bm{\tau}_g(\bm{q}) =  J^{T}(\bm{q})\bm{f}
    \label{eq:human_dynamics2}
\end{equation}

From equation (\ref{eq:human_dynamics2}), it can be seen that the gravity and body motion directly reduce the capacity to resist and apply external forces. 
%However, in many cases the gravitational and dynamical effects can be neglected $\bm{\tau}_g\!=\!\bm{\tau}_d\!=\!0$ considering only the influence of muscular forces to the Cartesian force generation
% \begin{equation}
%     \bm{\tau} =  J^{T}(\bm{q})\bm{f}
%     \label{eq:human_simple}
% \end{equation}

\subsection{Muscle model}
The joint torque $\bm{\tau}$ is generated by muscle-tendon units. One of the most well known muscle models was introduced by A. Hill \cite{hill1938heat} and later refined by F. Zajac \cite{zajac1989muscle}, where the muscle is approximated by a tensile force composed of an active and a passive component
\begin{equation}
    F_i = f^A_i(l_i,v_i)F_{iso,i} \alpha_i + \underbrace{f^P_{i}(l_i,v_i)F_{iso,i}}_{F_{p,i}}, \quad \alpha_i \in \left[0, 1\right]
\end{equation}

$f^A_{i}$ and $f^P_{i}$ are active and passive scaling functions depending on the muscle length $l_i$ and contraction/extension velocity $v_i$. $F_{iso,i}$ is the maximal isometric force the muscle can generate and $\alpha_i$ is the muscle activation level. For a set of $d$ muscles, with specified muscle lengths $\bm{l}$ and contraction velocities $\bm{v}$ the achievable set of muscle tensile forces $\bm{F} \in\mathbb{R}^d$ has a range
\begin{equation}
    \bm{F} \in \left[ \bm{F}_{p}, ~ \bm{F}_m\right]
    \label{eq:muslce_initial_range}
\end{equation}
where $\bm{F}_p$ is the vector of passive forces obtained for $\bm{\alpha}=0$ and $\bm{F}_m$ is the vector of maximal muscles forces obtained for $\bm{\alpha}=1$. 
% NR not necessarily the  maximal isometric force unless you are in the kineto-static case. More generally, FP is obtained for alpha_i = 0 and FM for alpha_i = 1.

The joint torque $\bm{\tau}$, generated by the muscle tensile force $\bm{F}$, can then be calculated using the Moment arm matrix \cite{pandy1994}   
\begin{equation}
    \bm{\tau} = -L^{T}(\bm{q})\bm{F} 
    \label{eq:muscle_torqe_gen}
\end{equation}
where $L(\bm{q})^T$ is the transpose of the moment arm matrix, which is defined as the muscle length Jacobian relating the space of joint and muscle length velocities  
\begin{equation}
    \dot{\bm{l}} = L(\bm{q}) \dot{\bm{q}},\quad L_{ij}=\dfrac{\partial l_i}{\partial q_j}
\end{equation}
Finally, the negative sign in equation (\ref{eq:muscle_torqe_gen}) makes the force applied in the length shortening direction of the muscle positive. 

Combining equations (\ref{eq:human_dynamics2}) and (\ref{eq:muscle_torqe_gen}), one obtains the relationship relating the $d$ muscle tensile forces $\bm{F}$ and  the $m$ dimensional Cartesian wrench $\bm{f}$
\begin{equation}
    J^{T}(\bm{q})\bm{f} = -L^{T}(\bm{q})\bm{F} - \bm{\tau}_d(\bm{q},\dot{\bm{q}},\ddot{\bm{q}})- \bm{\tau}_g(\bm{q})
\end{equation}

\subsection{Residual muscle forces}
For any given joint configuration $\bm{q}$, velocity $\dot{\bm{q}}$ and acceleration $\ddot{\bm{q}}$, the dynamics $\bm{\tau}_d$ and the gravity $\bm{\tau}_g$ torques are constants and can be considered as a bias. Their equivalent bias muscle force $\bm{F}_{b}$ can be defined as
\begin{equation}
    -L^{T}(\bm{q})\bm{F}_{b} = \bm{\tau}_d(\bm{q},\dot{\bm{q}},\ddot{\bm{q}}) + \bm{\tau}_g(\bm{q})
    \label{eq:torque_constraint}
\end{equation}
This force $\bm{F}_{b}$ can be assessed by finding the minimal muscle activation $\bm{\alpha}$ that generates the desired torque vector \cite{ANDERSON2001153}. This can be resolved using a quadratic problem formulation where the muscle activation vector norm $||\bm{\alpha}||$ is minimised, with equation (\ref{eq:torque_constraint}) as equality constraint 
\begin{equation}
\begin{aligned}
    \min_{\bm{F_b}} \quad &  \frac{1}{2}\bm{F}_{b}^TP\bm{F}_{b}\\
     \textrm{s.t.} \quad &  -L^T\bm{F}_{b}= \bm{\tau}_g + \bm{\tau}_d \\
          & \bm{F}_{p} \leq \bm{F}_{b} \leq \bm{F}_{m} \\
\end{aligned}
\end{equation}
where  $P$ is a diagonal matrix with elements on diagonal $ p_{ii}=1/(F_{m,i}\! -\! F_{p,i})^2 $. 

Furthermore, the bias muscle forces $\bm{F}_{b}$ can then be used to redefine the range of possible residual muscle forces $\bm{F}'$ from equation (\ref{eq:muslce_initial_range})

\begin{equation}
    \bm{F}' =  \bm{F} - \bm{F}_{b}, \quad \bm{F}' \in \left[\bm{0}, ~ \bm{F}_{m}\! -\! \bm{F}_{b} \right] 
\end{equation}

Finally, without loss of generality, the relationship of the muscle-tendon tension forces $\bm{F}$ and the Cartesian wrenches $\bm{f}$ can be redefined
\begin{equation}
    J^{T}(\bm{q})\bm{f} = -L^{T}(\bm{q})\bm{F}, \quad \bm{F} \!\in\! \left[\unum{\bm{F}},\,\onum{\bm{F}}\right]
\end{equation}
where the $\unum{\bm{F}}$ and $\onum{\bm{F}}$ are the minimal and maximal muscle-tendon tension forces $\bm{F}$.

\subsection{Wrench capacity polytope}\label{ch:wrench_feasabiliy_definition}

A general achievable set of muscle-tendon tensile forces  $\bm{F}\!\in\!\left[\unum{\bm{F}},\,\onum{\bm{F}}\right] $ forms a $d$-dimensional hyper-rectangle 
with side lengths equal to the ranges of each of the $d$ muscle forces. Using the moment arm matrix $L(\bm{q})$ and equation (\ref{eq:muscle_torqe_gen}) this hyper-rectangle can be projected into the $n$-dimensional space of joint torques $\bm{\tau}$, forming the convex polytope of achievable joint torques $\mathcal{P}_\tau$
\begin{equation}
    \mathcal{P}_\tau = \big\{ \bm{\tau}\! \in\! \mathbb{R}^{n} ~ | ~ \bm{\tau} =\! -L^{T}(\bm{q})\bm{F}, ~ \bm{F}\! \in\! \left[\unum{\bm{F}},\,\onum{\bm{F}}\right] \big\}
    \label{eq:torque_poly}
\end{equation}

Furthermore, the dual relationship of the $m$-dimensional Cartesian wrenches $\bm{f}$ and the generalised joint torques $\bm{\tau}$ is given through the Jacobian transpose matrix $J^{T}(\bm{q})\bm{f}\! = \!\bm{\tau}$ and defines the achievable Cartesian force polytope
\begin{equation}
    \mathcal{P}_f = \big\{ \bm{f}\! \in\! \mathbb{R}^{m} ~ | ~ J^{T}(\bm{q})\bm{f}\! =\! \bm{\tau}, ~ \bm{\tau}\! \in\! \mathcal{P}_\tau \big\}
    \label{eq:force_poly_1}
\end{equation}

Combining the definitions (\ref{eq:torque_poly}) and (\ref{eq:force_poly_1}) the implicit definition of $\mathcal{P}_f$ can be defined as
\begin{equation}
    \mathcal{P}_f = \left\{ \bm{f}\! \in\! \mathbb{R}^{m} ~ | ~ J^{T}(\bm{q})\bm{f}\! =\! -L^{T}(\bm{q})\bm{F}, \bm{F}\! \in\! \left[\unum{\bm{F}},\,\onum{\bm{F}}\right]\right\}
    \label{eq:full_polytope}
\end{equation}

\begin{figure}[!t]
    \centering
    \includegraphics[width=0.90\linewidth]{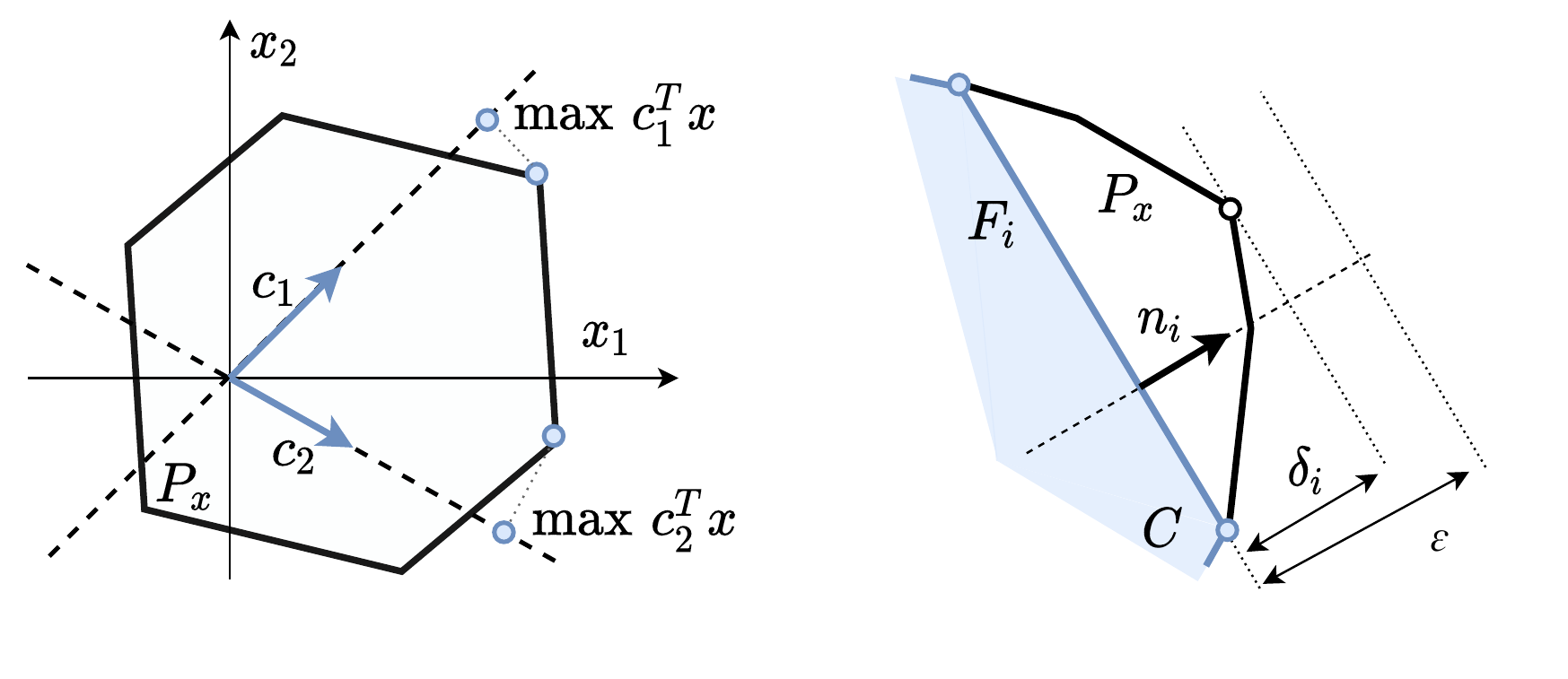}
    \caption{Left figure demonstrates the LP based vertex search by choosing two different vectors $\bm{c}$. Right figure exemplifies the polytope face condition from equations (\ref{eq:normal_coplanar_test}) and (\ref{eq:normal_distance}), where  face $\mathcal{F}_i$ of the convex-hull $\mathcal{C}$ is considered to be the face of $\mathcal{P}_x$, because $\delta_i\! < \!\varepsilon$.  }
    \label{fig:sample_problems}
\end{figure}

\section{Polytope evaluation formulation}
\label{ch:algorihtm}

Given the generic formulation of problem (\ref{eq:eq_general}) the set of all feasible $\bm{x}$ forms a convex polytope 
\begin{equation}
    \mathcal{P}_x = \big\{ \bm{x}\in \mathbb{R}^{m}\, |\,A\bm{x} = B\bm{y},\quad  \bm{y}\in \left[\unum{\bm{y}}, \onum{\bm{y}}\right]  \big\}
    \label{eq:poly_init_2}
\end{equation}

%In the \ref{ch:lp_adapt}, new LP formation is proposed adapting it to the implicit formulation (\ref{eq:full_polytope}) of the polytope $\mathcal{P}_f$. In the section \ref{ch:method}, the proposed algorithm overview is given, as well as the pseudo-code and the geometrical representation.

The polytope evaluation can either be done by finding the set of all the vertices ($\mathcal{V}$-rep) or by finding its half-space representation ($\mathcal{H}$-rep). Most of the commonly used polytope evaluation algorithms are very efficient for finding the exact solution to very specific problems: \cite{fukuda_dd}\cite{skuric:hal-02993408}\cite{sasaki2011vertex} calculate the $\mathcal{V}$-rep of $A\bm{x}\!=\!\bm{b},\bm{b}\!\in\![\bm{b}]$ and \cite{hyper_psm}\cite{dantzig1973fourier}\cite{jones2004equality} calculate the $\mathcal{H}$-rep of $\bm{x}\!=\! A\bm{y},\bm{y}\!\in\![\bm{y}]$. However since the number of faces and vertices of the polytopes augments exponentially with the size of the system, when the size of the problem becomes high, searching for the exact solution of the combined sub-problems becomes practically intractable \cite{Huynh2005PracticalIO}. 
%An example of such high-dimensional systems are the musculoskeletal models, where the number of muscles is usually large and even though the output space is low-dimensional Cartesian space the exact solution may take hours to calculate. 

To overcome this issue, various approximative approaches, such as the  \textit{Ray Shooting Method} (RSM)  \cite{agarwal1993ray} and the \textit{Convex-Hull Method} (CHM) \cite{lassez1992quantifier}, have been developed, reducing the complexity and improving the execution time. 
Although RSM algorithms are relatively simple to setup and have already been used for evaluating the capacity of musculoskeletal models \cite{carmichael2011Towards}, these algorithms do not provide any bound on their estimation error and highly rely on hand tuned initial parameters. 
CHM \cite{lassez1992quantifier} algorithms on the other hand, have shown to be very efficient while at the same time maintaining the bound of the user defined estimation error. CHM uses  \textit{linear-programming} (LP) repetitively to find new vertices and the convex-hull algorithm to group them to faces, augmenting the precision of the approximation successively. Additionally, CHM finds the $\mathcal{V}$-rep and the $\mathcal{H}$-rep of the polytope at the same time.  However in its standard formulation, the CHM is not suitable for the family of problems given with equation (\ref{eq:eq_general}). 

Therefore, Section \ref{ch:lp_adapt} introduces an LP formulation adapting the CHM approach to the implicit problem formulation (\ref{eq:eq_general}) and in Section \ref{ch:method}, the proposed algorithm overview is given, as well as the pseudo-code and the geometrical representation.

%The geometrical representation of solving the linear programming (LP) problem is finding a vertex of a polytope \cite{vajda_gass_1964}. However, in its implicit formation (\ref{eq:poly_init_2}) the polytope $\mathcal{P}_x$ is not suitable for the LP optimisation.

\subsection{Linear programming formulation}
\label{ch:lp_adapt}
From a geometrical point of view, solving a linear programming (LP) problem boils down to finding a vertex of a polytope \cite{vajda_gass_1964}. However, in order for the polytope $\mathcal{P}_x$ to be suitable for LP optimisation in the $m$-dimensional output space, its implicit formulation (\ref{eq:poly_init_2}) needs to be expressed in explicit form.

In the general case, $n\geq m$ and $A$ is not square. To find an explicit form of equation (\ref{eq:eq_general}), the pseudo-inverse $A^+$ can be used as a solution only if $B\bm{y}$ belongs to the \textit{image} of $A$ \cite{klema_singular_1980}. Therefore, equation (\ref{eq:eq_general}) can be transformed to 
\begin{equation}
    \bm{x} = A^{+} B \bm{y}, \quad B\bm{y} \in \mathcal{I}m(A)
    \label{eq:eq_general_red}
\end{equation} 
Using the \textit{Singular value decomposition} \cite{klema_singular_1980} of  $A = U\Sigma V^T$, an equality constraint can be devised for $B\bm{y}$ to belong to the \textit{image} of $A$
\begin{equation}
    U_2^TB\bm{y} = \bm{0}
    \label{eq:nullspace_gen}
\end{equation}
where rotation matrix $U \in \mathbb{R}^{n\times n }$ is separated into  $U_1\in \mathbb{R}^{n\times m}$, a projector to the  \textit{image} of $A$, and $U_2\in \mathbb{R}^{n\times(n-m)}$, a projector to its \textit{null-space}.

Finally, combining equations (\ref{eq:nullspace_gen}), (\ref{eq:eq_general_red}) and (\ref{eq:eq_general}), one can create a linear program \cite{vajda_gass_1964} capable of reaching all  vertices of the polytope $\mathcal{P}_x$
\begin{equation}
\begin{aligned}
    \max_{\bm{y}} \quad &  \bm{c}^TA^{+}B\bm{y} \\
     \textrm{s.t.} \quad &  U_2^TB\bm{y} = \bm{0} \\
          & \unum{\bm{y}} \leq \bm{y} \leq \onum{\bm{y}} \\
\end{aligned}
\label{eq:lin_prog}
\end{equation}
by appropriately choosing the projection $\bm{c}$, as demonstrated on figure \ref{fig:sample_problems}.

\subsection{Proposed algorithm overview}
\label{ch:method}

Given the polytope definition (\ref{eq:eq_general}) and (\ref{eq:poly_init_2}) and the LP problem defined in (\ref{eq:lin_prog}), the proposed method provides a structured way to choose vectors $\bm{c}$ to find all the vertices ( $\mathcal{V}$-rep ) and facets ( $\mathcal{H}$-rep ) of the polytope $\mathcal{P}_x$. 
In order to do so, the method leverages the iterative convex hull $\mathcal{C}$ calculation, where each newly found vertex extends $\mathcal{C}$. The normal vectors of the faces of the newly obtained $\mathcal{C}$ are used to decide for the new vectors $\bm{c}$ to use in the LP (\ref{eq:lin_prog}). This iterative process is performed until some specified level of accuracy is reached.

\begin{figure}[!t]
    \centering
    \includegraphics[width=\linewidth]{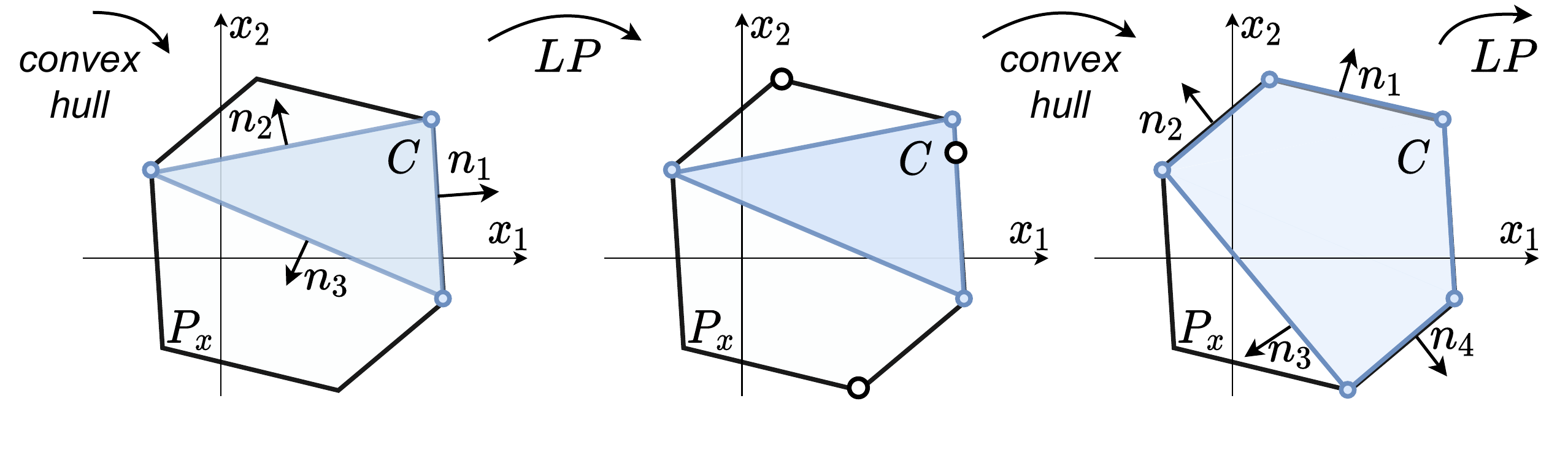}
    \caption{This figure shows the procedure of successive approximation of the polytope $\mathcal{P}_x$ using the proposed CHM algorithm. The face normal vectors $\bm{n}_i$ of the convex-hull $\mathcal{C}$ are used with LP (\ref{eq:lin_prog}) to find new vertices which are then used to update the convex-hull $\mathcal{C}$ and furthermore improve the approximation of the polytope. }
    \label{fig:algo_example}
\end{figure}

More precisely, the first step of the algorithm constructs an initial set of vertices $\bm{x_v}$ and an initial convex hull. In a $m$-dimensional space, the minimal number of points to create a volume is $m\!+\!1$. Since $\bm{x}$, defined by relation (\ref{eq:eq_general_red}), can be expressed as a linear combination of the $m$ base vectors $v_i\in V$ of $A$, $$\bm{x} = a_0 \bm{v}_0 + \dots +a_m \bm{v}_m$$ a way to get the initial set of vertices is to solve LP (\ref{eq:lin_prog}) that minimises ($\bm{c} = -\bm{v}_i$) and maximises ($\bm{c} = \bm{v}_i$) the projection of the polytope $\mathcal{P}_x$ on the axes of each one of the base vectors $\bm{v}_i$. Once the initial set of vertices $\bm{x}_v$ is found, the initial convex hull $\mathcal{C}$ can be calculated. Its centroid position is given by $\bm{x}_c\!=\!\frac{1}{N}\!\sum\bm{x}_{v,i}$. 

The next step of the algorithm is to iterate over all the faces $\mathcal{F}_i$ of the convex hull $\mathcal{C}$. For each new face $\mathcal{F}_i$, the normal $\bm{n}_i$ is found and used as a candidate $\bm{c}_i$ in the direction pointing out of polytope $\mathcal{P}_x$.
%all the vertices $\bm{x}_{\mathcal{F}j}\! =\! %\bm{x}_v\!\in\!\mathcal{F}_i$. Then the $\bm{c}_i$ is defined %as normal $\bm{n}_i$ in the direction pointing out of the polytope 5$\mathcal{P}_x$.
The normal vector direction is verified by projecting the vector going from the centroid $\bm{x}_c$ to any vertex $\bm{x}_{j}$ of face $\mathcal{F}_i$, onto the normal $\bm{n}_i$ and verifying if the scalar product is positive or negative.
\begin{equation}
    \bm{c}_i = \begin{cases}
  \bm{n}_i, & \text{if } \bm{n}_i^T(\bm{x}_{j} - \bm{x}_c) \geq 0, \\
  -\bm{n}_i, & \text{otherwise}.
\end{cases} 
\label{eq:normal_condition}
\end{equation}

\begin{figure*}[!t]
    \centering
    \includegraphics[width=0.9\textwidth]{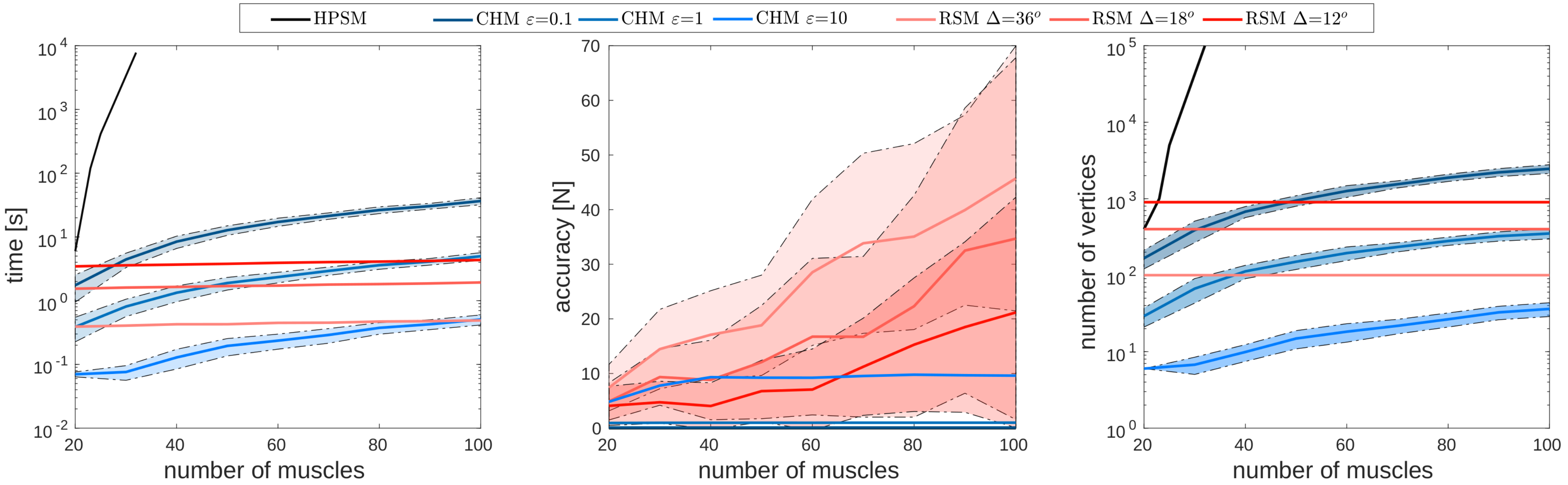}
    \caption{Figure presenting the performance analysis results for three different algorithms: HPSM, RSM and proposed CHM with respect to different number of muscles. Left figure shows the evolution of the execution time in the logarithmic scale. Middle figure shows the evolution of the underestimation error, the maximal distance in between the polytope $\mathcal{P}_x$ and the acquired polytope, calculated as max$\{|\delta_{i}|\}$. Right figure shows the evolution of the number of vertices found in logarithmic scale. All the plots show the averaged results over 100 algorithm runs, where each run corresponds to one randomly generated model.}
    \label{fig:performance_results}
\end{figure*}

Solving equation (\ref{eq:lin_prog}) with $\bm{c}_i$, the obtained $\bm{x}_i$ can be either a vertex of the polytope $\mathcal{P}_x$ or be coplanar with face $\mathcal{F}_i$. To determine if  $\bm{x}_i$ is coplanar with the face, a simple check can be devised, which verifies if the orthogonal (normal) distance from $\bm{x}_i$ to any vertex $\bm{x}_{j}$ of face $\mathcal{F}_i$
\begin{equation}
    \delta_i = \bm{n}_i^T(\bm{x}_{j} - \bm{x}_i)
\label{eq:normal_distance}
\end{equation}
is within a certain user defined accuracy $\varepsilon$:
\begin{equation}
    \bm{x}_i = \begin{cases}
   \text{vertex}, & \text{if }  |\delta_i| \geq \varepsilon, \\
    \text{on the face}, & \text{otherwise}.
\end{cases} 
\label{eq:normal_coplanar_test}
\end{equation}
Figure \ref{fig:sample_problems} provides a graphical interpretation of this check and of the values $\delta_i$ and $\varepsilon$.

If $\bm{x}_i$ belongs to face $\mathcal{F}_i$ of the convex-hull $\mathcal{C}$ then  $\mathcal{F}_i$ is considered to be a face of the polytope $\mathcal{P}_x$. In that case the algorithm updates the $\mathcal{H}$-rep of $\mathcal{P}_x$ 
\begin{equation}
    H \leftarrow \begin{bmatrix} H \\ \bm{n}_i^T\end{bmatrix}, \quad \bm{d} \leftarrow \begin{bmatrix}  \bm{d} \\ \bm{n}^T_i \bm{x}_{j} \end{bmatrix}
\label{eq:h_rep}
\end{equation}
where $\bm{n}_i$ is the face normal vector and $\bm{n}_i^T \bm{x}_{j}$ is the orthogonal distance from the origin to the face $\mathcal{F}_i$. On the other hand, if $\bm{x}_i$ is a vertex of the polytope it is appended to the $\mathcal{V}$-rep list $\bm{x}_v \leftarrow [\bm{x}_v, ~\bm{x}_i]$.

Once all the faces of the convex hull $\mathcal{C}$ are evaluated, the convex hull is updated using the new vertex list $\bm{x}_v$.
If the vertex list $\bm{x}_v$ has no new elements or, in other words, if the maximal distance $\delta_i$ for all the faces $\mathcal{F}_i$ is lower than the user defined accuracy $\varepsilon$  
\begin{equation}
    \max\{|\delta_{i}|\} \leq \varepsilon
\end{equation}
the procedure is completed. Otherwise the procedure restarts by iterating over all the new faces $\mathcal{F}_i$ of the updated convex hull $\mathcal{C}$. 

A pseudo-code of one example implementation of the proposed algorithm can be found in Algorithm \ref{alg:main_algo}. Furthermore, figure \ref{fig:algo_example} visually demonstrates several iterations of the algorithm for the 2-dimensional polytope $\mathcal{P}_x$ example. The code of the proposed algorithm used in this paper is publicly available\footnote{ \url{https://gitlab.inria.fr/askuric/human\_wrench\_capacity}}.

\begin{algorithm}[!h]
\caption{Proposed CHM algorithm pseudo-code}
\begin{algorithmic}
\REQUIRE $A$,$B$, $\unum{\bm{y}}$, $\onum{\bm{y}}$, $\varepsilon$
\STATE $U, \Sigma, V^T \leftarrow svd(A)$ 

\STATE init $\mathcal{H}$-rep: $H,\bm{d} \leftarrow [\,]$ and $\mathcal{V}$-rep: $\bm{x}_v\leftarrow [\,]$
\STATE \textbf{for all}   {$\bm{v}_i$ in $V$} \textbf{do}

\hspace{0.3cm} $\bm{y}_{i}\leftarrow$ LP (Eq. \ref{eq:lin_prog}) with $\bm{c}\! =\! \bm{v}_i$ and $\bm{c}\! =\! -\bm{v}_i$

\hspace{0.3cm} $\bm{x}_{v} \leftarrow [\bm{x}_{v}, ~ A^+B\bm{y}_i]$

\REPEAT

\STATE calculate the convex hull $\mathcal{C}$ of $\bm{x}_{v}$
\STATE calculate the centroid $\bm{x}_{c} = \frac{1}{N}\sum_i\bm{x}_{v,i}$

\FORALL{ new face $\mathcal{F}_i$ in $\mathcal{C}$ } 
\STATE find normal $\bm{n}_i$ and one vertex $\bm{x}_{j}$ of $\mathcal{F}_i$

\STATE $\bm{y}_{i}\leftarrow$ LP ( Eq. \ref{eq:lin_prog} ) with  $\bm{c}\! =\! \bm{c}_i$ ( Eq. \ref{eq:normal_condition} )

\STATE $\bm{x}_{i} = A^+B\bm{y}_i$ 

\STATE   calculate $\delta_i$ (Eq.\ref{eq:normal_distance})

\STATE  \textbf{if}   $ |\delta_i| \leq \varepsilon$ \textbf{then} 

\hspace{0.4cm}  new face: update $\mathcal{H}$-rep  ( Eq. \ref{eq:h_rep} )

\STATE \textbf{else}

\hspace{0.4cm} new vertex: update $\mathcal{V}$-rep $\bm{x}_{v} \!\leftarrow\! [\bm{x}_{v},~ \bm{x}_i ]$ 

\ENDFOR
\UNTIL{{ $\max\{|\delta_{i}|\}\leq \varepsilon$}}
\RETURN $\bm{x}_v$, $H$, $\bm{d}$ 
\end{algorithmic}
\label{alg:main_algo}
\end{algorithm}

\subsection{Feasible wrench polytope}
The application of the proposed method to compute the feasible  Cartesian wrench polytope $\mathcal{P}_f$ is direct. $\bm{x}$ is chosen as the Cartesian wrench $\bm{f}$, $\bm{y}$ with the range $\bm{y}\! \in\!\left[\unum{\bm{y}},\onum{\bm{y}}\right]$ is the vector of muscle-tendon tensile forces $\bm{F}\!\in\! \left[\unum{\bm{F}},\onum{\bm{F}}\right]$.  $A$ is the Jacobian transpose matrix $J^T(\bm{q})$ and $B$ is the moment arm matrix $-L^T(\bm{q})$. Finally, the accuracy $\varepsilon$ can be interpreted as the maximal underestimation error of the Cartesian wrench $\bm{f}$ allowable, representing the desired accuracy of the polytope estimation.

\section{Experiments and results}
\label{ch:results}

\subsection{Performance analysis}

To evaluate the performance of the proposed CHM algorithm, a comparative experiment is performed. The experiment consists in finding the $\mathcal{V}$-rep of the cartesian force $m=3$ polytope for a randomised mockup musculoskeletal model with $n\!=\!7$ degrees of freedom and a number of muscles ranging from $d$ = $20$ to $100$. 

Apart form the proposed method, two additional algorithms are tested. The first algorithm presents an exact approach. It combines the \textit{Hyper plane shifting method} (HPSM) \cite{hyper_psm}, to evaluate the $\mathcal{H}$-rep of the polytope $\mathcal{P}_\tau$, and the \textit{primal-dual} method introduced by Bremer et al. \cite{bremner_fukuda_marzetta_1998}, to find the $\mathcal{V}$-rep of $\mathcal{P}_f$. The second method is the RSM algorithm introduced by Carmichael et al. \cite{carmichael2011Towards} as an example of approximative algorithm used in the context of musculoskeletal models.

The RSM algorithm \cite{carmichael2011Towards} requires uniform sampling of the \textit{ray directions} in the 3D space. This sampling can be performed based on a two Euler angles parametrization. For this method, three levels of granularity are tested for each Euler angles: $\Delta=36^o$, $18^o$ and $12^o$, making for $100$, $400$ and $900$ tested \textit{ray directions} in 3D space. For the proposed CHM algorithm, three values of accuracy are tested $\varepsilon$= $0.1$, $1$ and $10$ N. All the algorithms are implemented in Matlab and run on a 1.90GHz Intel i7-8650U processor.

Results of the experiments, averaged over 100 algorithm runs, are shown on figure \ref{fig:performance_results}. The results confirm that the exact approach using the HPSM method has an execution time exponentially related to the number of muscles. Even for only 30 muscles it already takes more than 4.5hours to calculate, therefore this method is not tested on more than 30 muscles, where it finds over $10^4$ vertices.  

The RSM algorithm shows nearly constant time of execution for the full range of tested muscles and the constant number of vertices found, which is expected. The results show however, that for low muscle numbers $d\!<$25, when using a fine granularity $\Delta$=$12^o$ the RSM algorithm finds more vertices than the exact solution found by the HPSM. Furthermore, the middle plot of figure \ref{fig:performance_results} shows that the RMS estimation error increases considerably with the number of muscles $d$, followed with very high variance. These results confirm that, due to the fact that the polytope shape is not spherical, covering uniformly the space ray directions will necessarily lead RSM based algorithms to estimate certain areas of the polytope better than the others.

The graphs show that the proposed CHM method's execution time depends near-linearly of the number of muscles considered $d$ and the estimation error bound parameter $\varepsilon$. The accuracy graph, shown in the middle plot, shows that the proposed CHM algorithm is capable of limiting the estimation error of the polytope evaluation under desired value $\varepsilon$ regardless of the number of muscles. Furthermore, considering the vertex number found by the algorithms, it can be seen that the number of vertices has a nearly-linear relationship with the number of muscles $d$ and the variable $\varepsilon$. 

The demonstrated efficiency of the proposed CHM algorithm opens many doors for its applications in real-time systems, providing the user with an easy to understand trade-off between speed and accuracy.

\subsection{Collaborative object carrying}
In this experiment a human operator and a \textit{Franka Emika Panda} robot are jointly carrying an object with mass of $m_o\!=\!7$ kilograms, where each one carries a part of the total weight. 
$$
f_h + f_r = m_og = G_o
$$

The human operator is navigating the object through the common workspace, passing through 4 via-points, indicated visually with numbered markers on the table. The via-points are set on the corners of a square with 30cm side length.

\begin{figure}[!t]
    \centering
    \includegraphics[width=\linewidth]{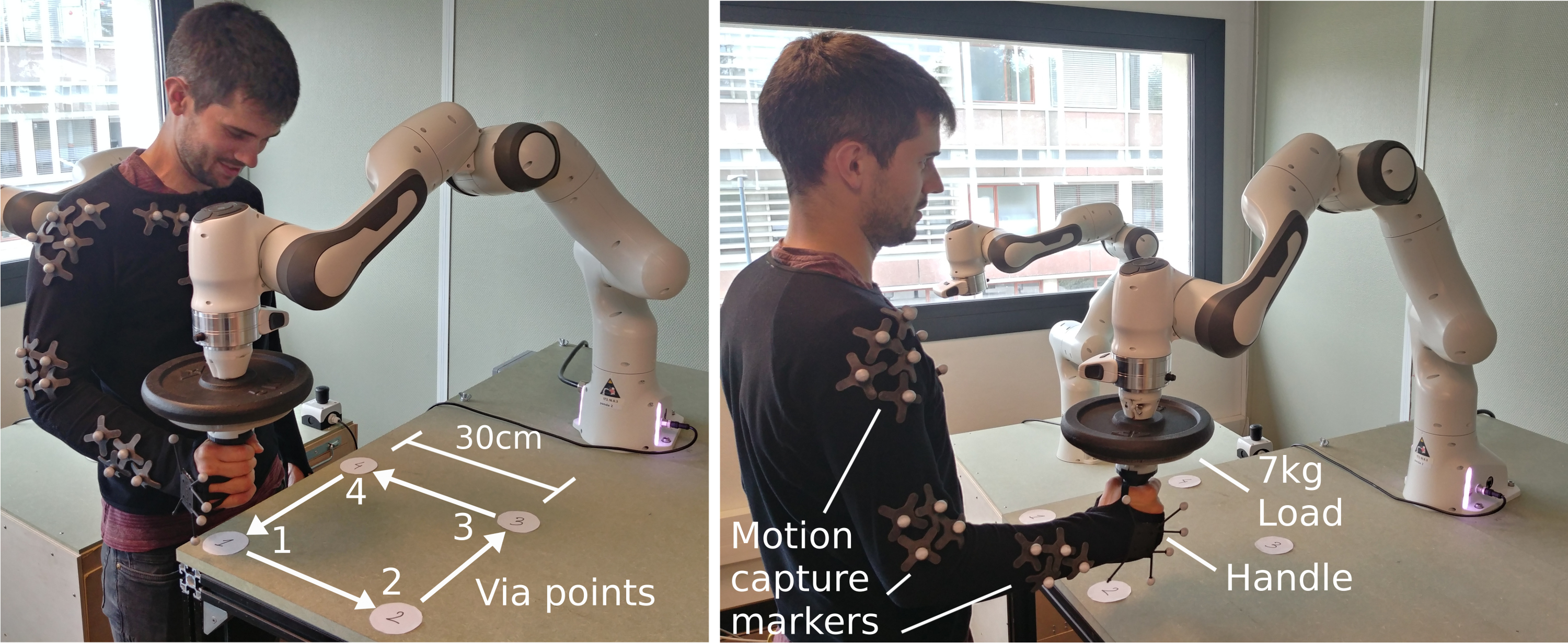}
    \caption{These figures show the experimental setup, where the human and robot jointly carry a 7 kg object. The robot end-effector is fixed to the object and the human operator is holding the object by the handle. The motion capture system is used to acquire the pose of the human arm. The via-points are visually indicated to the human operator with numbered stickers on the table placed at the corners of a 30 cm square. The left image is taken during the execution of the experiment 1 and the right one during the experiment 2.}
    \label{fig:experiment}
\end{figure}

The human upper limb configuration is inferred in real-time using an \textit{Optitrack} motion capture system. The full musculoskeletal model analysis of the human arm is developed using the efficient C++ library \textit{pyomeca biorbd} \cite{Michaud2021}. The musculoskemetal model used in this experiment is the 50 muscles, 7 degrees of freedom model MOBL-ARMS \cite{saul2015benchmarking}\cite{holzbaur2005model}. Based on this model and the proposed CHM algorithm, the carrying capacity of the human arm is calculated, in real-time, as the maximal force the human arm can generate in the $z$-axis (vertical) direction, forming a 1D polytope. The accuracy is set to $\varepsilon\!=\!0.1$N.

The robot is controlled following the \textit{Assist-As-Needed} (AAN)\cite{carmichael2013admittance} paradigm, ensuring that the human operator's relative load is kept constant with respect to its real-time carrying capacity. The fixed ratio value chosen for the experiments is 30\% of the human's carrying capacity.
$$
f_h = 0.3 f_{h,max}, \quad f_r = G_o - f_h
$$

The robot is controlled using the direct force control approach
$$
\bm{\tau}_r = J_{ee}^T(\bm{q}) f_r + \bm{\tau}_{g,r}(\bm{q})
$$
where $J_{ee}$ is the end-effector Jacobian, $\bm{\tau}_{g,r}$ is the joint torque vector necessary to compensate the robot's gravity and $\bm{\tau}_r$ are the joint torques applied to the robot. 
The carrying capacity of the robot is calculated \cite{skuric:hal-02993408} in the real-time, making sure that its capacity is not exceeded as well. 

All the algorithms are implemented in Python. The robot control interface is implemented using the \textit{Robot operating system} (ROS) and runs on a computer with 1.90GHz Intel i7-8650U processor.

Two experiments are conducted, as shown on figure \ref{fig:experiment}, in which the human operator follows the same sequence of via-points from two different positions in space, equally distant from the via-points but rotated by 90° degrees. Apart from the change in the operators position, no other changes are introduced.

\begin{figure}[!t]
    \centering
    \includegraphics[width=\linewidth]{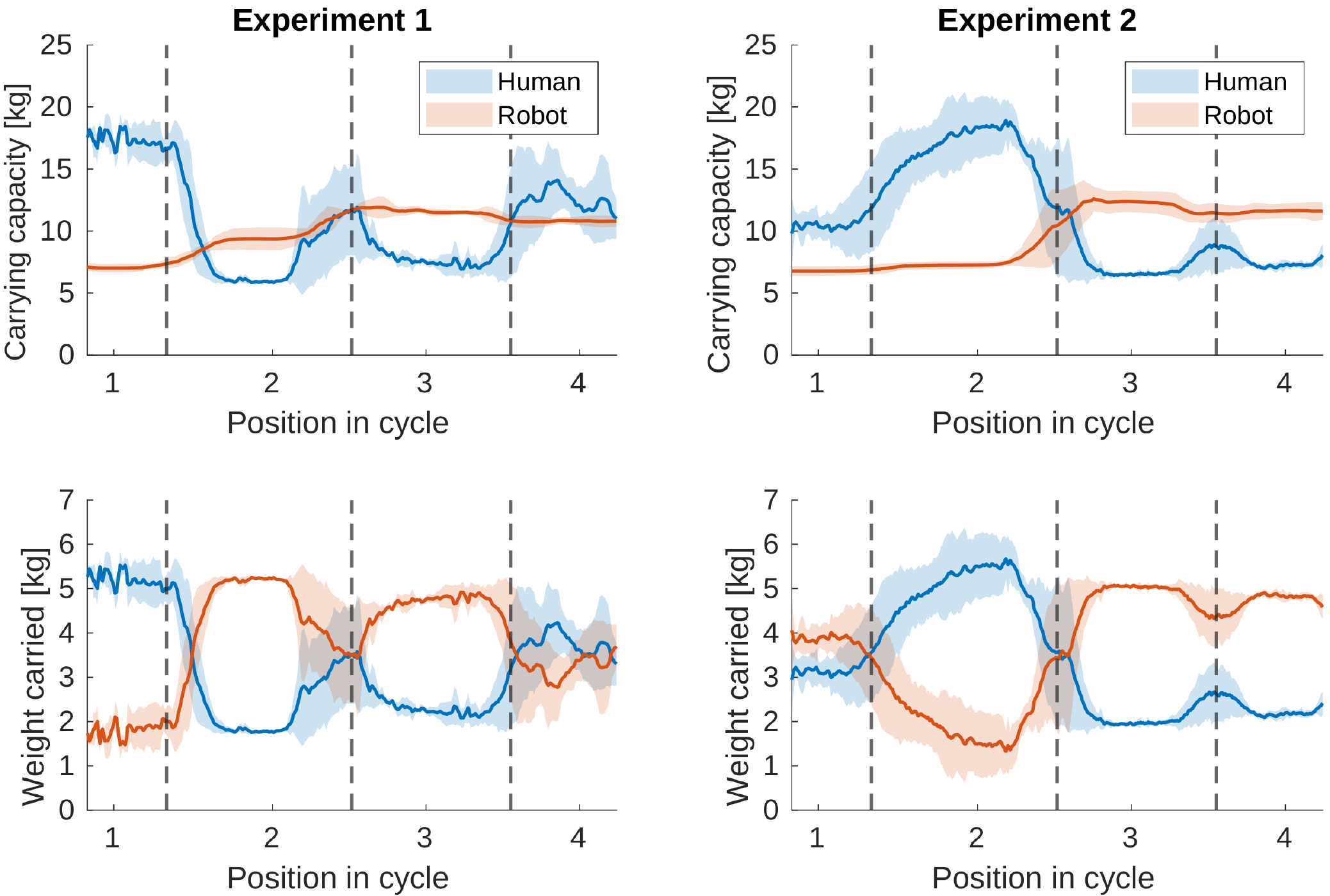}
    \caption{This figure shows the evolution of the carrying capacity and weight carried by the operator and the robot over the course of the via-point cycle, for two experiments. The plots show mean values and the variances of the curves calculated over 10 successive cycles. The graph regions belonging to the different via-points are separated by vertical lines.  All the values are expressed in kilograms for easier readability.}
    \label{fig:experiment_results}
\end{figure}

Figure \ref{fig:experiment_results} shows the graphs of the carrying capacity and the weight carried by the human and the robot for the two experiments. From the figure it can be seen that, as expected, the evolution of robot's carrying capacity is very similar in both experiments, this is due to the fact that the via-points are fixed to the same location in space with respect to the robot base for both experiments. The human carrying capacity in both experiments peaks around 20 kilograms demonstrating that the human operator's force capacity is not negligible with respect to the robot's, however its large variability, from 20kg to less than 5kg, emphasizes the importance of measuring it in real-time.    

Figure \ref{fig:experiment_results} further shows that even though the human's carrying capacity profile changes significantly from one experiment to the other, by calculating the operator's carrying capacity in real-time, the robot is successfully able to adapt its assistance degree and ensure that the operator's relative load does not surpass the desired level of 30\%.
To demonstrate this adaptability even further, in addition to  experiments 1 and 2 , a third experiment is included in the video attachment, where the human operator moves freely in space, without \textit{a priori} defined trajectory. 

\section{Conclusion}

A new generic iterative convex hull based algorithm for evaluating the polytope associated to the class of problems $A\bm{x}\!=\!B\bm{y}$ is introduced. The efficiency of the method is demonstrated on the evaluation of the feasible wrench polytope of human musculoskeletal models. In that context, the execution time of the proposed method is shown to be nearly-linear with respect to the number of muscles in the model as well as to the user defined accuracy. 

The potential of the method to be used for collaborative robot control is demonstrated in a collaborative carrying experiment, where the human operator and a robot jointly carry a 7kg object. The experiment shows that by combining the real-time human's capacity evaluation with a simple control law, a highly flexible collaboration scenario can be created. 

The proposed algorithm opens many possibilities in the domain of the on-line evaluation of human and robot capabilities and, consequently for the adaptive control of robots interacting with humans in dynamic contexts. Yet, some applicative challenges remain open. They both relate to the ability to evaluate human posture in real time in a minimally intrusive way and to the individual scaling and calibration of the used musculoskeletal models. They also relate to the inclusion of capability alteration models related to fatigue but also cognitive factors. This potential increase in models complexity advocates for systematic experimental validations of the results they lead to \cite{biomechanics1010008}.

% \section{Discussion}
% \todo[inline]{ maybe musculoskeletal models, calibration + scaling + precision}
%\addtolength{\textheight}{-6cm}   % This command serves to balance the column lengths

%%%%%%%%%%%%%%%%%%%%%%%%%%%%%%%%%%%%%%%%%%%%%%%%%%%%%%%%%%%%%%%%%%%%%%%%%%%%%%%%

%%%%%%%%%%%%%%%%%%%%%%%%%%%%%%%%%%%%%%%%%%%%%%%%%%%%%%%%%%%%%%%%%%%%%%%%%%%%%%%%

%%%%%%%%%%%%%%%%%%%%%%%%%%%%%%%%%%%%%%%%%%%%%%%%%%%%%%%%%%%%%%%%%%%%%%%%%%%%%%%%
%%%%%

\section*{Acknowledgment}
This work has been funded by the  BPI France Lichie project.

\newpage
\bibliographystyle{ieeetr} % <-- or: "plain"
\bibliography{ICRA2022}

\end{document}